%% file: main.tex
\documentclass[10pt,twocolumn,letterpaper]{article}
\input{constants}
\input{cvpr_header}
\input{preamble}

\unless\ifarxiv \myexternaldocument{_supplementary} \fi

\title{6DOPE-GS: Online 6D Object Pose Estimation using Gaussian Splatting}

\author{Yufeng Jin$^{1,2}$, Vignesh Prasad$^{1}$, Snehal Jauhri$^{1}$, Mathias Franzius$^{2}$, Georgia Chalvatzaki$^{1,3}$ \\
$^{1}$Computer Science Department, Technische Universit\"at Darmstadt, Germany \\ $^{2}$Honda Research Institute Europe GmbH, Offenbach, Germany $^{3}$Hessian.AI, Darmstadt, Germany \\ \{\texttt{yufeng.jin, vignesh.prasad, snehal.jauhri\}@tu-darmstadt.de}, \\
\texttt{georgia.chalvatzaki@tu-darmstadt.de}
}

\let\oldtwocolumn\twocolumn
\renewcommand\twocolumn[1][]{%
    \oldtwocolumn[{#1}{
    \begin{center}
        \vspace{-0.8cm}
        \includegraphics[width=1.0\textwidth]{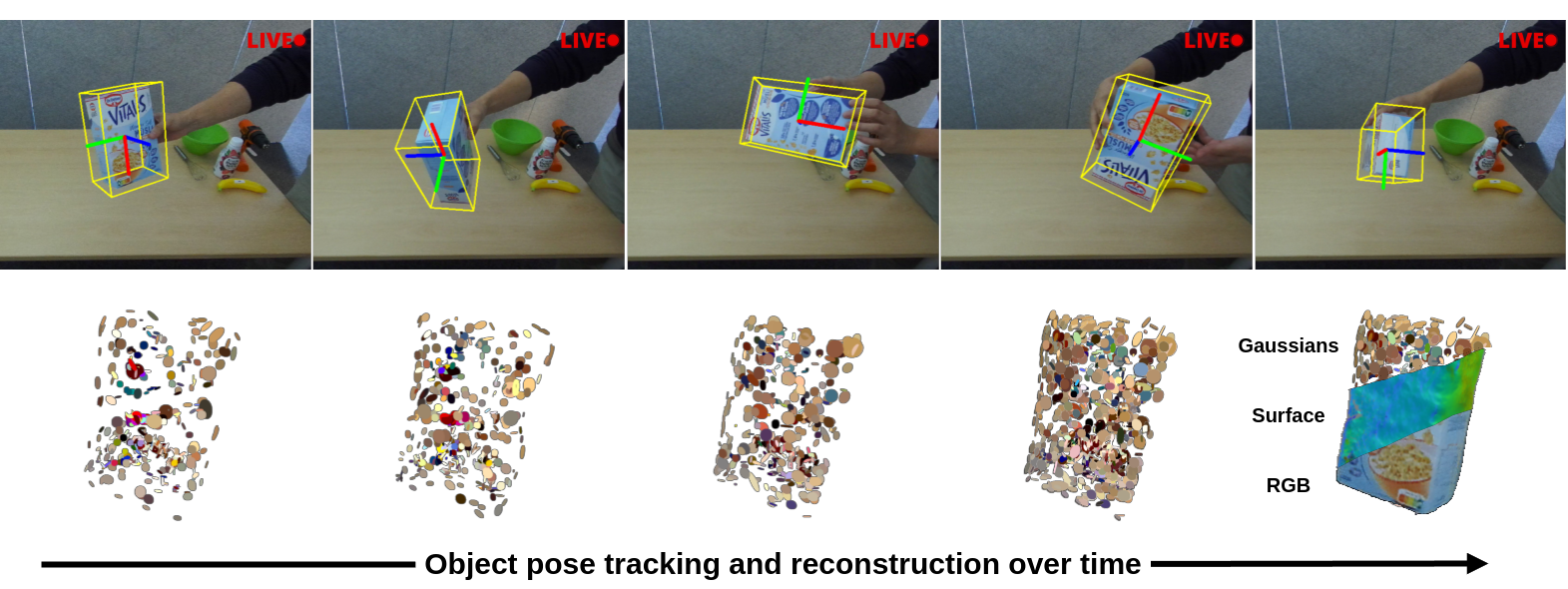}
        \captionof{figure}{Demonstrating live object pose tracking and reconstruction of a test object in the real-world using 6DOPE-GS: a novel method for joint 6D object pose estimation and reconstruction using Gaussian Splatting. \textbf{Top:} 6D pose estimates of the object over time, \textbf{Bottom:} Example reconstruction over time with 2D Gaussian disks used to render the surface and appearance of the object. Our method enables live pose tracking and Gaussian Splat reconstruction of dynamic objects at 3.5Hz.}
        \label{fig:teaser}
        \vspace{0.2cm}
        \end{center}
    }]
}
\begin{document}
\maketitle

\input{00_abstract}
\input{01_intro}

\input{02_related}

\input{03_method}
\input{04_experiments}
\input{05_conclusion}

{\small
\bibliographystyle{ieeenat_fullname}
\bibliography{references}
}

\end{document}

%% file: constants.tex
\def\paperID{10013}
\def\confName{ICCV}
\def\confYear{2025}

\newif\ifreview
\newif\ifarxiv
\newif\ifcamera
\newif\ifrebuttal

\reviewfalse     
\arxivtrue     
\camerafalse     
\rebuttalfalse   

%% file: cvpr_header.tex
\ifreview \usepackage[review]{cvpr} \fi
\ifarxiv \usepackage[pagenumbers]{cvpr} \fi
\ifrebuttal \usepackage[rebuttal]{cvpr} \fi
\ifcamera \usepackage{cvpr} \fi

\input{_macros}  

\usepackage{xr-hyper}

\makeatletter
\newcommand*{\addFileDependency}[1]{
  \typeout{(#1)}
  \@addtofilelist{#1}
  \IfFileExists{#1}{}{\typeout{No file #1.}}
}

\makeatother
\newcommand*{\myexternaldocument}[1]{
    \externaldocument{#1}
    \addFileDependency{#1.tex}
    \addFileDependency{#1.aux}
}

\definecolor{cvprblue}{rgb}{0.21,0.49,0.74}
\usepackage[pagebackref,breaklinks,colorlinks,allcolors=cvprblue]{hyperref}
\usepackage[capitalize]{cleveref}
\crefname{section}{Sec.}{Secs.}
\crefname{table}{Table}{Tables}
\crefname{figure}{Fig.}{Figs.}

\ifarxiv \crefname{appendix}{App.}{Apps.}
\else \crefname{appendix}{Suppl.}{Suppls.} \fi

%% file: _macros.tex
\usepackage{graphicx}	
\usepackage{amsmath}	
\usepackage{amssymb}	
\usepackage{booktabs}
\usepackage{times}
\usepackage{microtype}
\usepackage{epsfig}
\usepackage{caption}
\usepackage{float}
\usepackage{placeins}
\usepackage{color, colortbl}
\usepackage{stfloats}
\usepackage{enumitem}
\usepackage{tabularx}
\usepackage{xstring}
\usepackage{multirow}
\usepackage{xspace}
\usepackage{url}
\usepackage{subcaption}
\usepackage{xcolor}
\usepackage[hang,flushmargin]{footmisc}

\ifcamera \usepackage[accsupp]{axessibility} \fi

% ...

\ifarxiv  \fi

\newcommand{\R}[1]{{%
    \textbf{%
        \ifstrequal{#1}{1}{\textcolor{red}{R#1}}{%
        \ifstrequal{#1}{2}{\textcolor{blue}{R#1}}{%
        \ifstrequal{#1}{3}{\textcolor{magenta}{R#1}}{%
        \ifstrequal{#1}{4}{\textcolor{teal}{R#1}}{%
                           \textcolor{cyan}{R#1}%
        }}}}%
    }%
}}

%% file: preamble.tex
\usepackage{multirow}
\usepackage{graphicx}
\usepackage{subcaption}
\usepackage{booktabs}


%% file: 00_abstract.tex
\begin{abstract}

Efficient and accurate object pose estimation is an essential component for modern vision systems in many applications such as Augmented Reality, autonomous driving, and robotics. While research in model-based 6D object pose estimation has delivered promising results, model-free methods are hindered by the high computational load in rendering and inferring consistent poses of arbitrary objects in a live RGB-D video stream. To address this issue, we present 6DOPE-GS, a novel method for online 6D object pose estimation \& tracking with a single RGB-D camera by effectively leveraging advances in Gaussian Splatting. Thanks to the fast differentiable rendering capabilities of Gaussian Splatting, 6DOPE-GS can simultaneously optimize for 6D object poses and 3D object reconstruction. To achieve the necessary efficiency and accuracy for live tracking, our method uses incremental 2D Gaussian Splatting with an intelligent dynamic keyframe selection procedure to achieve high spatial object coverage and prevent erroneous pose updates. We also propose an opacity statistic-based pruning mechanism for adaptive Gaussian density control, to ensure training stability and efficiency. We evaluate our method on the HO3D and YCBInEOAT datasets and show that 6DOPE-GS matches the performance of state-of-the-art baselines for model-free simultaneous 6D pose tracking and reconstruction while providing a 5$\times$ speedup. We also demonstrate the method's suitability for live, dynamic object tracking and reconstruction in a real-world setting.
\end{abstract}

%% file: 01_intro.tex

\section{Introduction}
\label{sec:intro}
Precise tracking and accurate reconstruction of objects allows capturing essential spatial and structural information, essential for downstream tasks such as robotic manipulation~\cite{deng2020self,stevvsic2020learning}, augmented reality~\cite{su2019deep,zhao2023augmented}, automation~\cite{kleeberger2019large,gorschluter2022survey}, etc. 
The majority of 6D object pose estimation and tracking methods, whether for seen or unseen objects, have primarily used model-based techniques. Several approaches~\cite{nguyenGigaPoseFastRobust2024,ornekFoundPoseUnseenObject2024,labbe2022megapose,linSAM6DSegmentAnything2024,wenFoundationPoseUnified6D2024} use CAD models rendered from various angles during training and perform feature matching at inference time for rapid pose estimation. Such model-based approaches augmented with synthetic training data~\cite{wenFoundationPoseUnified6D2024} have shown state-of-the-art performance instance-level pose estimation. However, doing so requires either a CAD model or a small set of reference images annotated with the object poses, which becomes tedious as the number of unseen objects increases. 

On the other hand, there has been exciting progress in zero-shot, model-free methods over the past few years~\cite{wenBundleTrack6DPose2021, wenBundleSDFNeural6DoF2023} which require no additional prior information, other than an object mask. BundleSDF~\cite{wenBundleSDFNeural6DoF2023} operates in a model-free manner by jointly optimizing a ``neural object field" and the object poses by learning a 3D Signed Distance Field representation while concurrently running a global pose graph optimization. However, despite reporting near real-time pose optimization capabilities ($\sim$10Hz)
, the neural object field training is far from real-time\footnote{\url{https://github.com/NVlabs/BundleSDF/issues/122}}, which limits the average tracking frequency to $\sim$0.4Hz. 
The significant computational overhead associated with training the neural object field hinders its applicability in live dynamic scenarios, where rapid pose updates are crucial.


To address this limitation, we leverage Gaussian Splatting~\cite{kerbl3DGaussianSplatting2023,huang2DGaussianSplatting2024} which offers significantly better computational efficiency for real-time applications. We propose a novel method for online 6D object pose estimation through Gaussian Splatting, ``6DOPE-GS", that enables model-free, live object tracking and reconstruction. Building upon recent advances in using Gaussian Splatting for SLAM~\cite{keetha2024splatam}, 6DOPE-GS jointly optimizes object poses from observed keyframes and reconstructs a 3D object model on the fly using incremental 2D Gaussian Splatting~\cite{huang2DGaussianSplatting2024}. We propose several algorithmic enhancements to attain the required accuracy, efficiency, and training stability for live reconstruction and tracking. For accuracy, our method uses a novel dynamic keyframe selection mechanism to prioritize spatial coverage of the object and reconstruction confidence-based filtering to exclude keyframes with erroneous pose estimates. To maintain training stability and efficiency, we propose an adaptive Gaussian density control mechanism based on the opacity statistics of the Gaussians. 
Our contributions provide a significant speed-up in object pose estimation and tracking while maintaining high accuracy. In particular, we evaluate 6DOPE-GS on the HO3D and YCBInEOAT datasets and observe that it matches the state-of-the-art performance of competitive baselines while providing a 5$\times$ speedup. We also demonstrate the live, dynamic object tracking and reconstruction ability of 6DOPE-GS in a real-world setting.  To the best of our knowledge, we are the first method to perform joint object tracking and Gaussian Splat reconstruction live at 3.5Hz from a single RGB-D camera.


Our contributions are as follows:
\begin{itemize}
    \item We propose a novel method that effectively leverages 2D Gaussian Splatting for efficient and accurate model-free 6D object pose estimation and reconstruction.
    \item We leverage the computationally efficient differentiable rendering of Gaussian Splatting to jointly optimize a 2D Gaussian Splatting-based ``Gaussian Object Field" along with an object-centric pose graph of observed keyframes, that provides accurate, refined keyframe pose updates.
    \item We propose a dynamic keyframe selection approach based on the spatial coverage of the set of keyframes and a reconstruction confidence-based filtering mechanism to exclude keyframes with erroneous pose estimates.
    \item We incorporate a novel adaptive Gaussian density control mechanism based on opacity percentiles to filter out ``unimportant" Gaussians, thereby improving training stability and computational efficiency.
\end{itemize}

\begin{figure*}[ht!]
    \centering
    \includegraphics[width=0.9\textwidth]{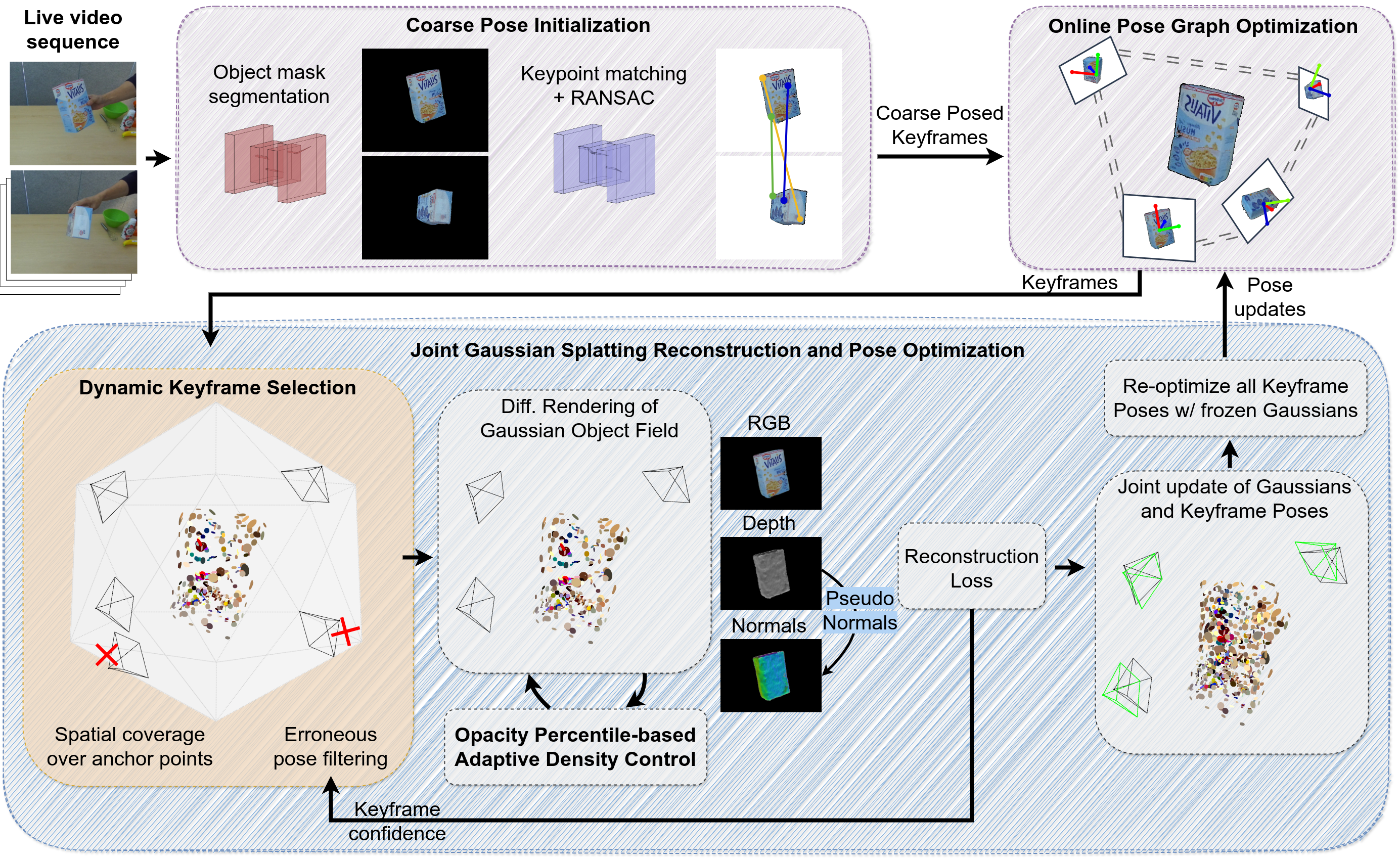}
    \caption{Overview of our approach: 6DOPE-GS. Given a live input RGB-D video stream, we obtain object segmentation masks using SAM2~\cite{raviSAM2Segment} on the incoming video frames. We then use LoFTR~\cite{sunLoFTRDetectorFreeLocal2021}, a transformer-based feature matching approach, to obtain pairwise correspondences between multiple views. We initialize a set of ``keyframes" based on the density of matched features, for which we establish initial coarse pose estimates using RANSAC. To obtain refined pose updates for the keyframes, we use a 2D Gaussian Splatting-based ``Gaussian Object Field" that is jointly optimized with the keyframe poses in a concurrent thread. We filter out erroneous keyframes for accurate pose refinement updates using a novel dynamic keyframe selection mechanism based on spatial coverage and reconstruction confidence. Moreover, we incorporate an opacity percentile-based adaptive density control mechanism to prune out inconsequential Gaussians, thus improving training stability and efficiency. Once the Gaussian Object Field is updated, it is temporarily frozen and the poses of keyframes that were filtered out are also updated. The object pose estimate at each timestep is then obtained by performing an online pose graph optimization using the incoming keyframe with the current set of keyframes.}
    \label{fig:pipeline}
\end{figure*}

%% file: 02_related.tex
\section{Related Work}

\subsection{Object Pose Estimation and Tracking}

Instance-level 6D object pose estimation typically requires object CAD models and/or pretraining~\cite{hePVN3DDeepPointWise2020, heFFB6DFullFlow2021, labbecosypose2020, wangDenseFusion6DObject2019, pengPVNetPixelWiseVoting2019, linSAM6DSegmentAnything2024,wangNormalizedObjectCoordinate2019,wenSe3TrackNetDatadriven6D2020,wang6PACKCategorylevel6D2019,chen2021sgpa}. Such instance-level methods can be further categorized into correspondence-based~\cite{radbb82017, tekinRealTimeSeamlessSingle2018, pavlakos6DoFObjectPose2017}, template-based~\cite{sundermeyerimplicit2018, dengPoseRBPFRaoBlackwellized2021}, voting-based~\cite{liukdfnet2021, hePVN3DDeepPointWise2020, nguyenGigaPoseFastRobust2024, heFFB6DFullFlow2021, wangDenseFusion6DObject2019}, and regression-based~\cite{gao6d2020, husingle2020} methods. For better generalization, some approaches use an object CAD model only at inference time~\cite{ornekFoundPoseUnseenObject2024, labbe2022megapose, shugurovosop2022}. Other methods~\cite{liuGen6DGeneralizableModelFree2023, parklatentfusion2020, heOnePoseKeypointFreeOneShot, heFS6DFewShot6D2022, sunonepose2022, wenFoundationPoseUnified6D2024, chen2024zeropose} relax this assumption by utilizing only few reference images of the object instead of the CAD model.

BundleTrack~\cite{wenBundleTrack6DPose2021} enables near real-time~($\sim$10Hz), model-free tracking with a SLAM-style approach. It uses keyframe point correspondences for coarse pose initialization with RANSAC, followed by object-centric pose graph optimization for refined estimates. BundleSDF~\cite{wenBundleSDFNeural6DoF2023} extends this by jointly performing pose tracking and object reconstruction through a neural object field, achieving state-of-the-art results in model-free settings. However, the neural field training is slow and computationally demanding ($\sim$6.7s per training round~\cite{wenBundleSDFNeural6DoF2023}), limiting its real-time applicability. We address this key limitation by leveraging the efficiency of Gaussian Splatting for joint object reconstruction and pose refinement, enabling effective live tracking.


\subsection{3D Reconstruction}

3D Reconstruction is a well-studied problem in Photogrammetry. Structure from Motion (SfM)~\cite{ozyecsil2017survey} is a commonly used approach to estimate camera poses and a sparse 3D structure from images without prior pose knowledge in an offline manner. Multi-View Stereo approaches (MVS)~\cite{furukawa2015multi,wang2021multi} build upon such pose estimates to refine a dense 3D reconstruction. For enabling more real-time reconstruction and pose tracking, Simultaneous Localization and Mapping (SLAM) methods~\cite{kazerouni2022survey,macario2022comprehensive,chen2022overview} approach the problem by jointly optimizing the camera poses and the environment reconstruction. Emerging methods that leverage neural representations have enhanced the fidelity of 3D reconstructions~\cite{wang2024dust3r,zhang2024monst3r,wang20243d,ren2023deepsfm}. Along similar lines, the use of Neural Radiance Fields (NeRFs)~\cite{mildenhall2021nerf} and Signed Distance Fields (SDFs)~\cite{ortiz2022iSDF,chibane2020neural,park2019deepsdf,mescheder2019occupancy}, with their volumetric rendering approach provide highly photorealistic reconstructions.

Gaussian Splatting~\cite{kerbl3DGaussianSplatting2023} is a particle-based alternative that models scene density with Gaussian distributions and achieves significantly faster rendering speeds with similar levels of photorealism by using rasterization of explicit Gaussian particles, thereby avoiding the ray-marching steps used by volumetric rendering methods. Recently, 2D Gaussian Splatting~\cite{huang2DGaussianSplatting2024} has improved the surface rendering capabilities of Gaussian Splatting by optimizing oriented planar 2D Gaussian disks close to the scene surfaces. However, all these methods still depend on pre-established camera poses. Coming from a SLAM perspective, recent approaches explore jointly optimizing camera pose estimates as well as the map reconstruction that uses Gaussian Splatting~\cite{keetha2024splatam,yan2024gs,matsuki2024gaussian,luiten2024dynamic}. In this work, we propose a novel method that extends scene-level approaches to object-level tracking and reconstruction by leveraging the SLAM-inspired capabilities for object tracking~\cite{wenBundleTrack6DPose2021,wenBundleSDFNeural6DoF2023} and Gaussian Splatting~\cite{keetha2024splatam,yan2024gs,matsuki2024gaussian} with the precise surface rendering capabilities offered by 2D Gaussian Splatting~\cite{huang2DGaussianSplatting2024}.

%% file: 03_method.tex
\section{Method}
\label{sec:approach}
We introduce a novel method for real-time 6D Object Pose Estimation using the representation capabilities of 2D Gaussian Splatting. Fig.~\ref{fig:pipeline} presents a schematic overview of our approach.
To accurately track the 6DoF pose of an object captured by a single RGB-D camera, we start by segmenting the object in the first frame using SAM2~\cite{raviSAM2Segment} to ensure precise object segmentation throughout the video sequence. With the object segmented across multiple frames, we apply LoFTR~\cite{sunLoFTRDetectorFreeLocal2021} to establish point correspondences, identifying keyframes for a Coarse Pose Initialization via Bundle Adjustment~\cite{wenBundleTrack6DPose2021} (Sec.~\ref{ssec:coarse-pose-initialization}). This initial set of coarsely estimated keyframes is then refined through a joint optimization with 2D Gaussians using differentiable rendering, yielding accurate pose corrections and an improved object model for the keyframes (Sec.~\ref{ssec:gaussian-object-field}). To improve the quality of the generated 3D model and to subsequently enable a more accurate pose refinement, we propose a dynamic keyframe selection technique for selecting the best keyframes for optimizing the 2D Gaussians based on their estimated spatial coverage around the object and their reconstruction accuracy (Sec.~\ref{ssec:dynamic-keyframe-selection}). During this phase, we iteratively employ a novel pruning/adaptive density control mechanism to stabilize the number of Gaussian particles required, to balance computational efficiency with reconstruction accuracy (Sec.~\ref{ssec:opacity-percentile-pruning}). Once the joint optimization converges, all the keyframe poses are subsequently optimized and help guide the Online Pose Graph Optimization (Sec.~\ref{ssec:pose-graph-optimization}) in continuously refining the object pose at each subsequent timestep for robust and precise tracking. 

\subsection{Coarse Pose Initialization}
\label{ssec:coarse-pose-initialization}
To enable real-time 6D pose tracking and reconstruction of arbitrary objects, we first use SAM2~\cite{raviSAM2Segment} for facilitating effective segmentation and tracking of the object in question. Specifically, we use a fixed-length window of past frames combined with prompted images as input. 
We then use LoFTR~\cite{sunLoFTRDetectorFreeLocal2021}, a transformer-based dense feature matcher, to estimate feature point correspondences between neighboring images. Using these matches, we compute a coarse pose estimate between pairs of RGB-D frames with nonlinear least-squares optimization~\cite{arun1987least} in a RANSAC fashion~\cite{fischler1981random}. 
Subsequently, a keyframe memory pool is created wherein if an incoming frame is deemed to be spatially diverse compared to the existing pool, it is added as a new keyframe. Further details regarding the feature matching and the keyframe memory pool initialization are in~\cite {wenBundleSDFNeural6DoF2023}.

\subsection{Gaussian Object Field}
\label{ssec:gaussian-object-field}
%

To build an internal model that captures the visual and geometric properties of the object in an efficient and accurate manner, we build a Gaussian Object Field using 2D Gaussian Splatting (2DGS)~\cite{huang2DGaussianSplatting2024} to achieve precise surface geometry reconstruction. Unlike 3D Gaussian Splatting (3DGS)~\cite{kerbl3DGaussianSplatting2023}, which primarily emphasizes on redering realistic visual effects, 2DGS ensures accurate geometric alignment by converting each Gaussian into a disk-like surfel. This surfel-based approach, combined with our novel dynamic keyframe selection (Sec.~\ref{ssec:dynamic-keyframe-selection}) and opacity quartile-based pruning (Sec.~\ref{ssec:opacity-percentile-pruning}), enables 2DGS to precisely model the rendered surface, thereby delivering reliable depth estimates and addressing the limitations observed in 3DGS. 



In 3DGS, a scene is represented as a set of 3D Gaussian particles, each of which represents a 3D distribution and is defined by its 3D centroid (mean)~$\boldsymbol{\mu}\in\mathbb{R}^3$ and a covariance matrix~$\boldsymbol{\Sigma}\in\mathbb{R}^{3\times3}$ which can be decomposed into a diagonalized scaling matrix~$\boldsymbol{S}=diag([s_x,s_y,s_z])$ and a rotation matrix $\boldsymbol{R}\in~SO(3)$~as~$\boldsymbol{\Sigma}~=~\boldsymbol{R}\boldsymbol{S}\boldsymbol{S}^{\top}\boldsymbol{R}^{\top}$, which denotes the volume (spread) of the particle in 3D space.
Along with the mean and covariance, each Gaussian is further characterized by spherical harmonic coefficients $c \in \mathbb{R}^k$ to represent view-dependent appearance, and an opacity value $\alpha~\in~[0,1]$. For rendering, each 3D Gaussian is converted to camera coordinates using the world-to-camera transformation matrix $\boldsymbol{W}$ and mapped to the image plane via a local affine transformation $\boldsymbol{J}$, $
\boldsymbol{\Sigma}' = \boldsymbol{J} \boldsymbol{W} \boldsymbol{\Sigma} \boldsymbol{W}^{\top} \boldsymbol{J}^{\top}$.
Once the 3D Gaussian is ``splatted" onto the image plane, excluding the third row and column of $\boldsymbol{\Sigma}'$ results in a 2D covariance matrix $\boldsymbol{\Sigma}^{2D}$ that represents a 2D Gaussian $G^{2D}$ in the image plane. 
The Gaussians are first ordered in ascending order based on their distance to the camera origin. Using volumetric rendering, we calculate the per-pixel color estimates $\hat{c}(\boldsymbol{p})$ of a pixel $\boldsymbol{p}=[u,v]^T$ as the $\alpha$-blending of $N$ ordered Gaussians from front to back along the view direction
\begin{equation}
\label{eq:gs-forward-rendering}
\hat{c}(\boldsymbol{p}) = \sum_{i \in N} c_i \alpha_i G^{2D}_i(\boldsymbol{p}) \prod_{j=1}^{i-1} (1 - \alpha_jG^{2D}_j(\boldsymbol{p})),
\end{equation}
where $\alpha_i$ and $c_i$ denote the opacity and the view-dependent appearance of the $i$th Gaussian, respectively. The depth image can be similarly rendered by replacing $c_i$ with the z-depth coordinate of the $i$th Gaussian in the camera frame. 

For 2DGS~\cite{huang2DGaussianSplatting2024}, the $z$-component of the scaling matrix is set to zero $\boldsymbol{S} = diag([s_u, s_v, 0])$ for each Gaussian, thereby collapsing the 3D volume into a set of 2D oriented planar Gaussian disks with two principal axes $\boldsymbol{t}_u$ and $\boldsymbol{t}_v$. The normal to the 2D Gaussian can then be defined as~$\boldsymbol{t}_w=\boldsymbol{t}_u\times\boldsymbol{t}_v$, which allows us to define the rotation matrix for the Gaussian particle as $\boldsymbol{R} = [\boldsymbol{t}_u, \boldsymbol{t}_v, \boldsymbol{t}_w]$. Moreover, along with photometric reconstruction, 2DGS additionally incorporates depth distortion and normal consistency to further enhance the quality
of the reconstructions. For further details regarding 2DGS, please refer to~\cite{huang2DGaussianSplatting2024}.

In our approach, along with optimizing the parameters of each 2D Gaussian, we aim to jointly refine the keyframe poses as well. We do so by propagating the gradients of losses through the projection operation of the 2D Gaussians onto the image plane of each keyframe, as done in~\cite{keetha2024splatam,yan2024gs,matsuki2024gaussian}. We use automatic differentiation via PyTorch~\cite{paszke2017automatic} for calculating the gradients and updating the keyframe poses. Further details are in Appendix.

\subsection{Dynamic Keyframe Selection for Gaussian Splatting Optimization}
\label{ssec:dynamic-keyframe-selection}
Once we obtain a coarse pose initialization of keyframes, we aim to construct a 2DGS model of the object. However, errors in the pose initialization can cause a divergence in the Gaussian Splatting optimization. Unlike BundleSDF’s ray-casting method~\cite{wenBundleSDFNeural6DoF2023}, which renders individual pixels, Gaussian Splatting uses tile-based rasterization, rendering entire images one at a time, thereby increasing the computational cost linearly as the number of keyframes increases. To mitigate these issues, we introduce a dynamic keyframe selection approach to filter out erroneous keyframes.

To acquire a reliable Gaussian Object Field, we strategically sparsely select keyframes to optimize keyframe poses and object Gaussians. We first establish a series of ``anchor points" at varying resolution levels, using the vertices and face centers of an icosahedron (as shown in~Fig.~\ref{fig:pipeline}-bottom left) to approximate evenly distributed points on a sphere centered on the object~\cite{saff1997distributing}. We then cluster the initial coarse keyframe pose estimates around these anchor points along the icosahedron. To maximize information from all viewpoints around the object, we select the keyframe with the largest object mask in the cluster of each anchor point, effectively training under sparse-view conditions with the aid of depth information~\cite{li2025geogaussian}. This ensures that we minimize instances where the object is largely occluded and consider views where we have better visibility of the object.

While jointly optimizing the 2D Gaussians and the selected keyframe poses, we further remove outliers with erroneous pose estimates based on the reconstruction error obtained during the 2D Gaussian optimization. This approach is necessary because reconstruction residuals can impede pose optimization during the joint optimization of 2D Gaussians and keyframe poses. Specifically, we estimate the median absolute deviation (MAD) of the reconstruction loss at each iteration, which represents the typical ``spread" around the median value, to identify and remove outlier views. The rationale for using MAD lies in its robustness; as a median-based metric, MAD is less influenced by extreme values than other measures, such as the mean or standard deviation, making it more reliable in the presence of outliers. Views with absolute deviations exceeding three times the MAD are classified as outliers.



\subsection{Opacity Percentile-based Adaptive Density Control}
\label{ssec:opacity-percentile-pruning}
During the optimization of the Gaussian Object Field, we perform periodic pruning and densification to maintain both the number and compactness of Gaussians. However, the vanilla Adaptive Density Control proposed in 3DGS has several limitations~\cite{buloRevisingDensificationGaussian2024}, since it demands considerable engineering work to adjust densification intervals and fine-tune opacity thresholds to stabilize training. 
Prior work~\cite{deng2023compressing} demonstrates that a gradual iterative pruning strategy can yield significantly sparser models while preserving high fidelity. Similarly, Fan et al.~\cite{fan2023lightgaussian} propose an importance weighting based on the scale percentile and the opacity of the Gaussians. However, they mainly focus on efficient compression of Gaussians. Inspired by~\cite{fan2023lightgaussian}, and since we have an object-centric focus, we limit the scale of the Gaussians and instead use a percentile-based pruning strategy based on opacity for stabilizing the number of Gaussians.

After a fixed number of optimization steps, we prune the Gaussians with opacity in the bottom 5th percentile until the opacity of the 95th percentile of the Gaussian particles exceeds a given threshold. This allows us to ensure that during the forward rendering (Eq.~\ref{eq:gs-forward-rendering}), a good number of high-quality Gaussian particles remain and those that are more inconsequential get pruned out. We empirically verify that our approach compared to naive absolute thresholding~\cite{kerbl3DGaussianSplatting2023}, improves the performance of our method. We further trigger splitting and cloning of the Gaussian particles when the positional gradient exceeds a predefined threshold, similar to~\cite{kerbl3DGaussianSplatting2023}. Notably, the variation of the positional gradient remains relatively stable and does not continuously increase during training. 
Once the optimization of the Gaussian Object Field converges, the poses of all the keyframes are refined using the reconstruction of the RGB, depth, and normals, by temporarily freezing the 2D Gaussians.


\subsection{Online Pose Graph Optimization}
\label{ssec:pose-graph-optimization}

When we receive the updated poses for the keyframes from the Gaussian Object Field, we establish a global object-centric coordinate system and a keyframe memory pool, which stores key correspondences. 
To balance computational efficiency with memory usage and reduce long-term tracking drift when a new frame is observed, a set of overlapping frames from the memory pool is selected for graph optimization based on the view frustum of the incoming frame. For each frame in this pool, we generate a point-normal map and compute the dot-product of the normals with the camera-ray direction of the new frame, to assess visibility. Frames are selected if the visibility ratio of the new incoming frame exceeds a defined threshold. We choose the best keyframes from the pool to construct the pose graph along with the incoming frame. We optimize the pose graph using pairwise geometric consistency by minimizing a dense pixel-wise re-projection error as in~\cite{wenBundleTrack6DPose2021}.

%% file: 04_experiments.tex
\section{Experiments}

\begin{figure*}[ht!]
    \centering
    \includegraphics[width=0.93\textwidth]{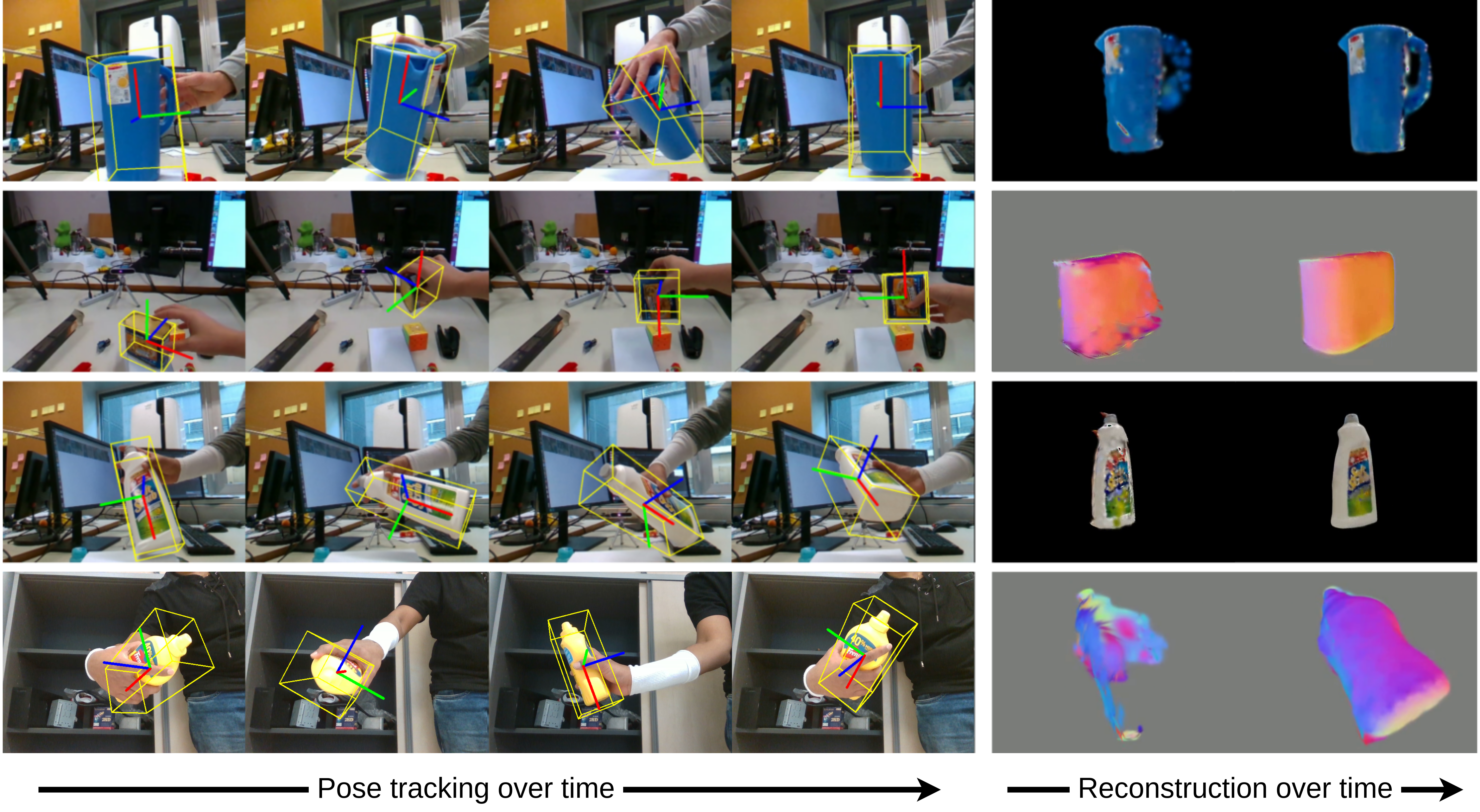}
    \caption{Qualitative results of our method, 6DOPE-GS, tested on video sequences from the HO3D dataset, namely AP13, MPM14, SB13, and SM1 (from top to bottom). \textbf{Left:} Our method tracks the 6D object pose over time with high accuracy, \textbf{Right:} 6DOPE-GS is effective at reconstructing both the appearance (rows 1 and 3) and surface geometry (rows 2 and 4) of the object over time. The first image shows the initial reconstruction at the beginning of the sequence, the second image shows the refined reconstruction over time.}
    \label{fig:qual_results}
\end{figure*}

\subsection{Datasets}

\subsubsection{YCBInEOAT Dataset} The YCBInEOAT dataset~\cite{wen2020se} offers ego-centric RGB-D video recordings of a dual-arm Yaskawa Motoman SDA10f robot manipulating YCB objects. Using an Azure Kinect camera positioned at mid-range, the dataset captures three types of manipulation tasks: single-arm pick-and-place, within-hand manipulation, and handoff between arms. In total, it includes 5 YCB objects~\cite{xiang2017posecnn} across 9 video sequences, amounting to 7,449 frames. Each frame is annotated with accurate 6D object poses, calibrated with the camera’s extrinsic parameters.

\subsubsection{HO3D Dataset} The HO3D dataset~\cite{hampali2020honnotate} features 27 multi-view (68 single-view) sequences showing hand-object interactions involving 10 YCB objects~\cite{xiang2017posecnn} with annotated 3D poses. The RGB-D video data, captured at close range with an Intel RealSense camera, provides detailed records of hand manipulations of objects. Ground-truth 3D poses are generated via multi-view registration, facilitating evaluations of pose, reconstruction, and texture accuracy. We use the latest version, HO3D\_v3, and conduct evaluations on the official test set, which comprises 4 objects and 13 sequences.


\subsection{Metrics \& Baselines}
We evaluate the performance of different methods based on three key metrics: 6-DoF object pose tracking accuracy, 3D reconstruction accuracy, and computational efficiency. Pose estimation accuracy is assessed using the area under the curve percentage for the ADD and ADD-S (ADD-Symmetric) metrics~\cite{heFS6DFewShot6D2022, xiang2017posecnn}. For object reconstruction, we measure the Chamfer distance between the reconstructed and ground-truth object meshes in object coordinates. Computational efficiency is evaluated based on the average processing time per frame.

We compare several SLAM-based approaches, including DROID-SLAM (RGBD)~\cite{teed2021droid}, NICE-SLAM~\cite{zhu2022nice}, KinectFusion~\cite{newcombe2011kinectfusion}, and SDF-2-SDF~\cite{slavcheva2016sdf}. We report the performance of other approaches listed on the leaderboard of~\cite{wenBundleSDFNeural6DoF2023}. We also evaluate the recent 3DGS-SLAM-based method MonoGS~\cite{matsukiGaussianSplattingSLAM2024}, along with BundleTrack~\cite{wenBundleTrack6DPose2021} and BundleSDF~\cite{wenBundleSDFNeural6DoF2023}, using their open-source implementations\footnote{\url{https://github.com/NVlabs/BundleSDF}}$^,$\footnote{\url{https://github.com/muskie82/MonoGS}} with optimized parameters. Each method takes as input an RGBD video and a per-frame mask indicating the object of interest. For a fair comparison, we evaluate MonoGS using only RGBD input.

\begin{table}[h!]
\resizebox{\columnwidth}{!}{%
\begin{tabular}{l||cc|c|c}
\hline
\multirow{2}{*}{Method} & \multicolumn{2}{c|}{Pose} & Reconstruction & Efficiency \\ \cline{2-5} 
 & ADD-S (\%) $\uparrow$ & ADD (\%) $\uparrow$ & CD (cm) $\downarrow$ & ATPF(s) $\downarrow$ \\ \hline
NICE-SLAM~\cite{zhu2022nice} & 23.41 & 12.70 & 6.13 & n.a.\\
SDF-2-SDF~\cite{slavcheva2016sdf} & 28.20 & 14.04 & 2.61 & n.a.\\
DROID-SLAM~\cite{teed2021droid} & 46.39 & 34.68 & 4.63 & n.a. \\
MaskFusion~\cite{runz2018maskfusion} & 41.88 & 35.07 & 2.34 & n.a.\\ \hline
MonoGS(RGB-D)~\cite{matsukiGaussianSplattingSLAM2024} & 20.16 & 15.32 & 2.43 & 0.29 \\
BundleTrack~\cite{wenBundleTrack6DPose2021} & 92.54 & 84.91 & - & \textbf{0.21} \\
BundleSDF~\cite{wenBundleSDFNeural6DoF2023} & 92.82 & 84.28 & 0.53 & 0.82 \\
\rowcolor[rgb]{ .988,  .894,  .839} 6DOPE-GS & \textbf{93.79} & \textbf{87.82} & \textbf{0.15} & 0.22 \\

\hline
\end{tabular}
}
\centering
\vspace{-0.3em}
\caption{Comparison on the YCBInEOAT Dataset. ADD and ADD-S are reported as AUC percentages (0 to 0.1m), and reconstruction accuracy is measured by Chamfer loss. ATPF is the average processing time per frame (n.a. indicates unavailable data).} 
\label{tab:ycbineoat-results}
\vspace{+0.3em}
\resizebox{\columnwidth}{!}{%
\begin{tabular}{l||cc|c|c}
\hline
\multirow{2}{*}{Method} & \multicolumn{2}{c|}{Pose} & Reconstruction & Efficiency \\ \cline{2-5} 
 & ADD-S (\%) $\uparrow$ & ADD (\%) $\uparrow$ & CD (cm) $\downarrow$ & ATPF(s) $\downarrow$ \\ 
NICE-SLAM~\cite{zhu2022nice} & 22.29 & 8.97 & 52.57 & n.a. \\
SDF-2-SDF~\cite{slavcheva2016sdf} & 35.88 & 16.08 & 9.65 & n.a.\\
KinectFusion~\cite{newcombe2011kinectfusion} & 25.81 & 16.54 & 15.49 & n.a.\\ 
DROID-SLAM~\cite{teed2021droid} & 64.64 & 33.36 & 30.84 & n.a. \\ \hline
MonoGS(RGB-D)~\cite{matsukiGaussianSplattingSLAM2024} & 2.81 & 1.82 & 22.09 & 0.36 \\
BundleTrack~\cite{wenBundleTrack6DPose2021} & 93.96 & 77.75 & - & 0.29\\
BundleSDF~\cite{wenBundleSDFNeural6DoF2023} & 94.86 & \textbf{89.56} & 0.58 & 2.10 \\

\rowcolor[rgb]{ .988,  .894,  .839} 6DOPE-GS & \textbf{95.07} & 84.33 & \textbf{0.41}  & \textbf{0.24} \\ \hline
\end{tabular}
}
\centering
\vspace{-0.3em}
\caption{Comparison on the HO3D Dataset. ADD and ADD-S are reported as AUC percentages (0 to 0.1m), and reconstruction accuracy is measured by Chamfer loss. ATPF is the average processing time per frame (n.a. indicates unavailable data). }
\label{tab:ho3d-results}
\vspace{-1.2em}
\end{table}

\subsection{Results}
As seen in Tables~\ref{tab:ycbineoat-results} and~\ref{tab:ho3d-results}, we outperform previous SLAM-based approaches as well as BundleTrack~\cite{wenBundleTrack6DPose2021}. In the YCBInEOAT dataset (Table~\ref{tab:ycbineoat-results}), where object motions are relatively smooth with less viewpoint diversity, most approaches perform similarly due to the absence of large occlusions or abrupt motion discontinuities that could lead to erroneous coarse pose initialization. 
We also compare our method to MonoGS~\cite{matsukiGaussianSplattingSLAM2024}, a state-of-the-art SLAM approach that employs Gaussian splitting for camera pose relocalization. On the YCBInEOAT dataset, MonoGS performs suboptimally due to limited geometric and textural supervision, leading to degraded object pose refinement, particularly in low-texture regions. On the HO3D dataset (Table~\ref{tab:ho3d-results}), which presents more challenging scenarios with complex hand-object interactions and rapid motion variations, all baselines struggle with accurate tracking due to accumulating errors. 
In contrast, our pose graph optimization and keyframe selection enhance robustness in the coarse pose initialization and pose tracking efficiency, which further results in superior reconstruction at a sub-centimeter level. 6DOPE-GS maintains a strong balance between pose accuracy and temporal efficiency, making it well-suited for real-time applications.

While 6DOPE-GS performs better than BundleSDF~\cite{wenBundleSDFNeural6DoF2023} in the ADD-symmetric score, we still fall behind in the absolute score on the more challenging HO3D dataset. 
One possible reason is the occlusions in HO3D, which reduce the supervisory signals needed to optimize Gaussian Particles and refine pose estimates. In contrast, BundleSDF uses mini-batch SDF rendering, allowing multiple correlated optimization steps and achieving better performance.
However, the high computational cost of BundleSDF limits its applicability for real-time scenarios. Our method, by contrast, provides a well-balanced trade-off between efficiency and accuracy, making it a more practical choice for real-world applications. Some examples of the precise pose tracking and accurate surface and appearance reconstruction of 6DOPE-GS are shown in Fig.~\ref{fig:qual_results}.

\begin{figure}[ht!]
    \centering
        \includegraphics[width=0.45\textwidth]{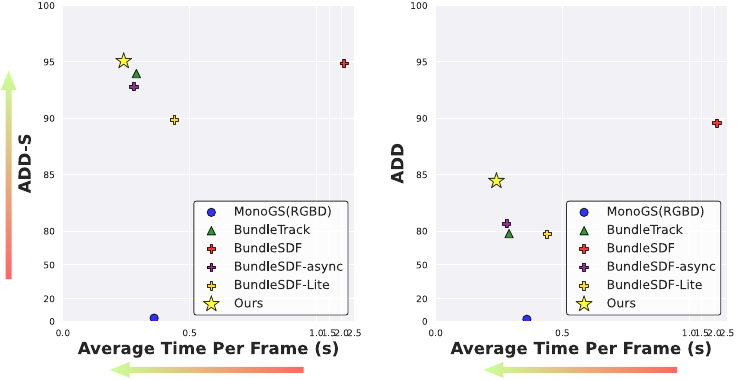}
        \caption{Comparison between temporal efficiency and performance for different approaches on the HO3D dataset. While BundleSDF achieves high performance, it comes at the cost of speed. On the other hand, 6DOPE-GS achieves a favorable tradeoff between speed and performance.}
    \label{fig:speed-performance-tradeoff}
    \vspace{-1em}
\end{figure}

\subsection{Temporal Efficiency}
\label{ssec:temporal-efficiency}
To evaluate the temporal efficiency of different approaches, we compare the tradeoff between the performance and the average processing time per frame for the different approaches on the HO3D dataset. We test the approaches on a desktop with a 12th Gen Intel(R) Core(TM) i9-12900KF CPU, 64GB RAM, equipped with an NVIDIA GeForce RTX 4090 GPU. We explore the performance of two more versions of BundleSDF~\cite{wenBundleSDFNeural6DoF2023} that reduce the processing time. In the first version, named ``BundleSDF-async", we deactivate the synchronization between the threads performing neural object field learning and the online pose graph optimization. While this introduces potential variability in pose accuracy, it reduces latency and achieves more real-time performance. In another version, ``BundleSDF-lite", we reduce the number of optimization steps for learning the neural object field, enabling faster synchronization between the threads.

From Fig.~\ref{fig:speed-performance-tradeoff}, we observe that the high pose tracking accuracy of BundleSDF~\cite{wenBundleSDFNeural6DoF2023} comes at a high computational cost. Since the pose tracking thread waits for the neural object field thread to converge and subsequently synchronize the object pose and reconstruction estimates, it requires an average processing time of 2.1 seconds. Surprisingly, BundleSDF-async (without the sync between the threads) achieves better performance than BundleSDF-lite even though BundleSDF-async runs the pose estimation without waiting for the neural object field. This highlights the dependence of the pose graph optimization on accurate keyframe poses. While the neural object field training in BundleSDF-async yields more accurate poses (although at a delayed timestep) than BundleSDF-lite, the pose estimation of the latter diverges given the premature termination of the optimization to achieve faster speeds. In contrast, 6DOPE-GS provides a balanced tradeoff between speed and accuracy. We achieve competitive performance without having to compromise on speed ($\sim$5$\times$ speedup over BundleSDF) as a result of the rapid convergence of the Gaussian Object Field optimization.

\subsection{Ablation Study}
We assessed our design choices on both the HO3D and YCBInEoat dataset, chosen for its variety of scenarios, the results of which are shown in Table~\ref{tab:ablation}. \textit{Ours (basic)} is a simplified version of 6DOPE-GS that naively uses all keyframes and employs vanilla adaptive density control. \textit{Ours w/o KF Selection} removes the dynamic keyframe selection strategy (Sec.~\ref{ssec:dynamic-keyframe-selection}) and performs joint optimization using all keyframes. \textit{Ours w/o Pruning} replaces the Opacity Percentile-based Adaptive Density Control~\ref{ssec:opacity-percentile-pruning} and employs vanilla adaptive density control~\cite{kerbl3DGaussianSplatting2023}. We also compare 2DGS and 3DGS representation,  \textit{Ours (3DGS)} replaces the 2D Gaussian representation with a 3D Gaussian representation for pose estimation and reconstruction.

Performance was reduced without dynamic keyframe selection (\textit{Ours w/o KF selection}) due to the retention of inaccurate pose estimates during training, which introduces residual errors in the reconstruction loss and hinders pose optimization. Applying the vanilla adaptive density control (\textit{Ours w/o Pruning}), where all Gaussians below a predefined threshold are removed, causes abrupt changes in the number of Gaussians. This results in significant rendering fluctuations, slowing the convergence of training. The pose accuracy and reconstruction quality of 3DGS (\textit{Ours (3DGS)}) are inferior to 2DGS. This can be attributed to the lack of regularization on the normal and depth in 3DGS, causing the Gaussians to deviate from the object surface and consequently degrading the reconstruction quality. We find that our approach with the proposed additions, namely Dynamic Keyframe Selection and the Opacity Percentile-based Adaptive Density Control performs the best among all.

\begin{table}[h!]
\centering
\resizebox{\columnwidth}{!}{
\begin{tabular}{l|c||cc|c}
\hline
\multirow{2}{*}{} & \multirow{2}{*}{Method} & \multicolumn{2}{c|}{Pose} & Reconstruction \\
\cline{3-5}
&  & ADD-S (\%) $\uparrow$ & ADD (\%) $\uparrow$ & CD (cm) $\downarrow$ \\
\hline
\multirow{5}{*}{\rotatebox[origin=c]{90}{HO3D}} 
& Ours (basic) & 93.52 & 80.25 & 0.44 \\
& Ours w/o KF Selection & 94.44 & 82.40 & 0.42 \\
& Ours w/o Pruning & 92.48 & 80.87 & 0.44 \\
& Ours (3DGS) & 92.51 & 79.49 & 0.47 \\
& Ours (final) & \textbf{95.07} & \textbf{84.33} & \textbf{0.41} \\
\hline
\multirow{5}{*}{\rotatebox[origin=c]{90}{YCBInEOAT}} 
& Ours (basic) & 92.74 & 85.15 & 0.22 \\
& Ours w/o KF Selection & 93.03 & 86.40 & 0.19 \\
& Ours w/o Pruning & 92.64 & 86.22 & 0.20 \\
& Ours (3DGS) & 91.18 & 85.29 & 0.41 \\
& Ours (final) & \textbf{93.79} & \textbf{87.82} & \textbf{0.15} \\ \hline
\end{tabular}
}
\caption{Ablation study of critical design choices}
\label{tab:ablation}
\vspace{-1em}
\end{table}

\subsection{Realtime Results}

\begin{figure}[h!]
    \centering
    \includegraphics[width=\linewidth]{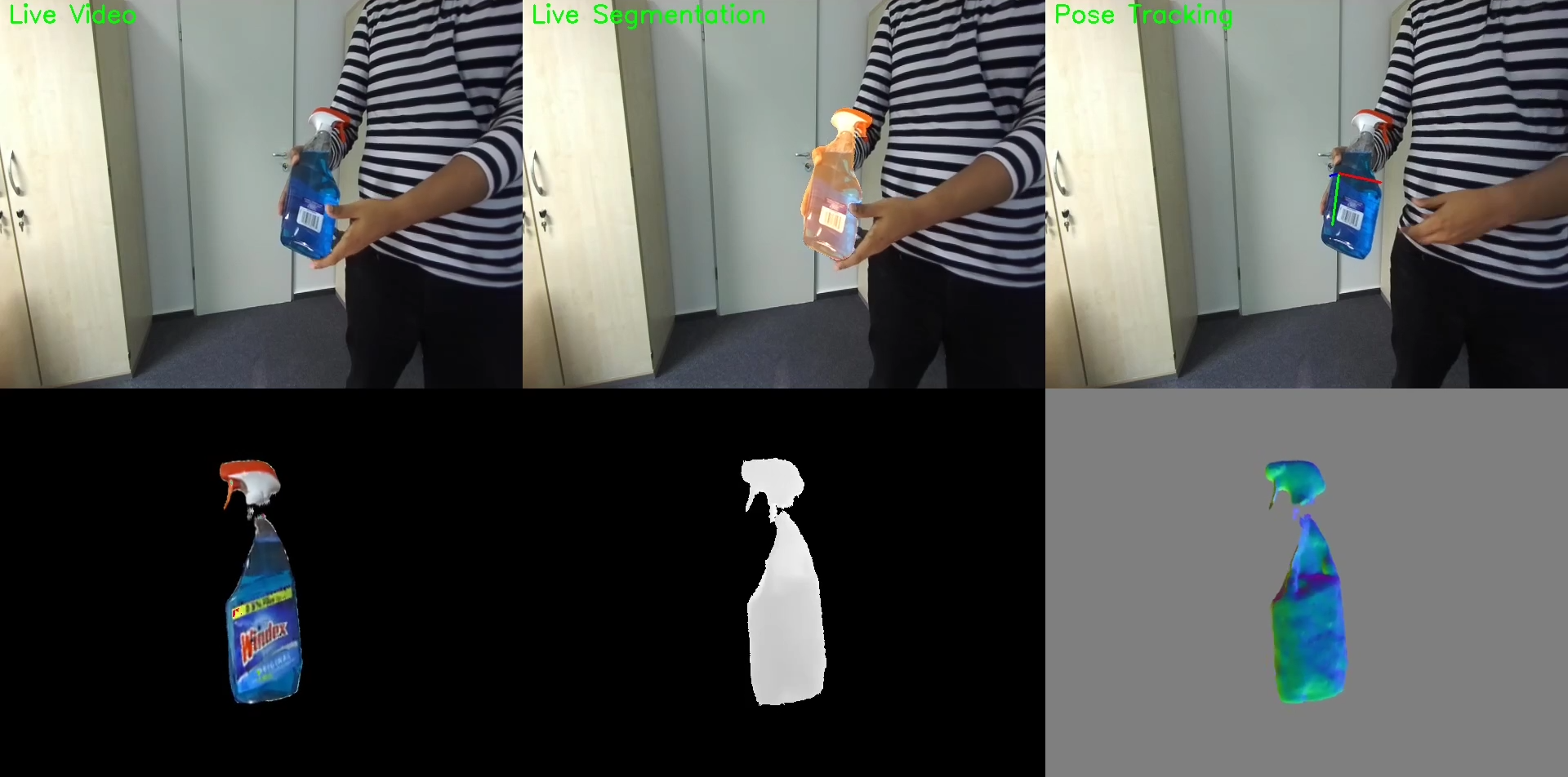}
    \caption{Example of real-time object tracking. \textbf{Top row:} Live video, object segmentation results, and pose tracking results. \textbf{Bottom row:} Rendered outputs, including color, depth, and surface normals derived from the Gaussian models.}
    \label{fig:realtime}
    \vspace{-1em}
\end{figure}

We utilized the ZED 2 camera operating in the standard depth sensing mode to maintain a balance between frame rate and accuracy. The camera captures video at a resolution of 1080p with a frame rate of 30 FPS. An initial mask for the target object was manually created through human annotation. The SAM2 system also operates at 28 FPS. Pose tracking, when running in visualization mode, achieves a processing frequency of 3-4 Hz, primarily due to the computational overhead introduced by the GUI and the rendering of Gaussian models in the background. Without the GUI, the system can operate at a slightly higher frequency of 4-5 Hz. The Gaussian model updates approximately every 8 seconds, as illustrated in Fig~\ref{fig:realtime}. For a more comprehensive understanding of the system's performance, we encourage readers to refer to the supplementary video provided.

%% file: 05_conclusion.tex
\section{Conclusion}
In this paper, we proposed ``6DOPE-GS", a novel method for model-free 6D object pose estimation and reconstruction that leveraged 2D Gaussian Splatting for jointly optimizing object pose estimates and 3D reconstruction in an iterative manner. Key to our method’s efficiency were a novel dynamic keyframe selection mechanism based on spatial coverage, as well as a confidence-based filtering mechanism to remove erroneous keyframes, followed by an opacity percentile-based adaptive density control for pruning out inconsequential Gaussians. These contributions enabled 6DOPE-GS to achieve competitive performance in a computationally efficient manner ($\sim$5$\times$ speedup), as validated on the HO3D and YCBInEOAT datasets, successfully capturing a practical balance of speed, accuracy, and stability for dynamic tracking scenarios in near real-time.

However, some shortcomings remain, which we aim to address in future work. Although Gaussian rasterization rendering is highly efficient and allows for rapid refinement of small translation and in-plane rotation errors, it may be less effective in gradient computations compared to differentiable ray casting used by neural radiance fields. To address this, we plan to investigate ray casting for rendering Gaussian representations~\cite{moenne20243d,guIRGSInterReflectiveGaussian2024}, which could improve both performance and computational efficiency. Another potential limitation is that the optimized 2D Gaussians are not directly integrated into the online pose graph optimization; instead, only the optimized poses are used. In future work, we will explore ways to more closely couple the trained object representation with the pose graph optimization.